\pdfoutput=1

\documentclass[11pt]{article}

\usepackage[preprint]{acl}

\usepackage{times}
\usepackage{latexsym}

\usepackage[T1]{fontenc}

\usepackage[utf8]{inputenc}

\usepackage{microtype}

\usepackage{inconsolata}

\usepackage{graphicx}

\usepackage{amsmath}
\usepackage{mathtools}
\usepackage{mathrsfs}
\usepackage{amsfonts}
\usepackage{bm}
\usepackage{bbm}
\usepackage{booktabs}
\usepackage{tabularx}
\usepackage{tablefootnote}
\usepackage{float}
\usepackage{xspace}
\usepackage{CJKutf8}
\usepackage{listings}
\usepackage{enumitem}
\usepackage[subrefformat=parens]{subcaption}

\newcommand{\mbrs}{\textsc{mbrs}\xspace}
\DeclareMathOperator*{\argmax}{argmax}

\DeclareMathOperator*{\expect}{\mathbb{E}}
\DeclareMathOperator{\supp}{Supp}

\newcommand{\MetricStyle}{\textsc}
\newcommand{\MetricBLEU}{\MetricStyle{Bleu}\xspace}
\newcommand{\MetricChrF}{\MetricStyle{chrF}\xspace}
\newcommand{\MetricCOMET}{\MetricStyle{Comet}\xspace}
\newcommand{\MetricBLEURT}{\MetricStyle{Bleurt}\xspace}
\newcommand{\MetricKiwi}{\MetricStyle{Kiwi}\xspace}
\newcommand{\MetricCometKiwi}{\MetricStyle{CometKiwi}\xspace}
\newcommand{\MetricTER}{\MetricStyle{Ter}\xspace}
\newcommand{\MetricXCOMET}{\MetricStyle{xComet}\xspace}

\newcommand{\regmark}{${}^{\text{\textregistered}}$}
\newcommand{\trademark}{${}^{\text{\texttrademark}}$}

\makeatletter
\def\lst@CJK@XXX#1#2#3{%
  \lst@ifletter\else\lst@OutputOther\fi
  \lst@whitespacefalse
  \advance\lst@length\@ne
  \lst@AppendOther{\let\lst@hss\relax\csname CJK@\number`#1\endcsname{`#2}{`#3}}%
}%
\lst@AddToHook{SelectCharTable}{%
  \let\CJK@XXX\lst@CJK@XXX
}
\makeatother

\title{\textsc{mbrs}: A Library for Minimum Bayes Risk Decoding}

\author{
  \textbf{Hiroyuki Deguchi}, \
  \textbf{Yusuke Sakai}, \
  \textbf{Hidetaka Kamigaito}, \
  \textbf{Taro Watanabe}
\\
  Nara Institute of Science and Technology (NAIST)
\\
  \texttt{\{deguchi.hiroyuki.db0, sakai.yusuke.sr9, kamigaito.h, taro\}@is.naist.jp}
}

\begin{document}
\maketitle
\begin{abstract}
Minimum Bayes risk (MBR) decoding is a decision rule of text generation tasks that outperforms conventional maximum a posterior (MAP) decoding using beam search by selecting high-quality outputs based on a utility function rather than those with high-probability.
Typically, it finds the most suitable hypothesis from the set of hypotheses under the sampled pseudo-references.
\mbrs is a library of MBR decoding, which can flexibly combine various metrics, alternative expectation estimations, and algorithmic variants.
It is designed with a focus on speed measurement and calling count of code blocks, transparency, reproducibility, and extensibility, which are essential for researchers and developers.
We published our \mbrs as an MIT-licensed open-source project, and the code is available on
GitHub\footnote{GitHub: \url{https://github.com/naist-nlp/mbrs}\label{footer:github}}${}^{,}$\footnote{Read the Docs: \url{https://mbrs.readthedocs.io/en/latest/index.html}}${}^{,}$\footnote{
YouTube: \url{https://youtu.be/4qeHpg4PTn0}
}${}^{,}$\footnote{HP: \url{https://naist-nlp.github.io/mbrs-web}}.

\end{abstract}

\section{Introduction}
\label{sec:intro}

Text generation has now become a central topic in natural language processing owing to the success of language models.
A typical text generation model generates the high-probability text using beam search and other search algorithms, which relies on maximum a posteriori (MAP) decoding; however, recent studies have demonstrated that high-probability texts are not always high-quality.
This phenomenon is known as \textit{beam search curse} and the models sometimes generate pathologically broken texts, e.g., empty sequences, $n$-gram repetitions, and copies of the inputs~\citep{koehn-knowles-2017-six,ott-etal-2018-analyzing,eikema-aziz-2020-map}.

Minimum Bayes risk (MBR) decoding addresses the problems of MAP decoding by determining outputs according to the decision rule based on quality or preference rather than probability.
Its effectiveness has been confirmed in statistical automatic speech recognition~\citep{goel-and-byrne-2000-minimum} and statistical machine translation~\cite{kumar-byrne-2004-minimum}, and it has been applied to neural text generation, especially neural machine translation~\citep{eikema-aziz-2020-map,muller-sennrich-2021-understanding}.
Although many variants of MBR decoding have been proposed, there is no common library where we can try the latest algorithms and compare them systematically.
It is clearly essential for both researchers and developers to establish a library for the further improvement in MBR decoding.

\begin{figure}[t]
    \centering
    \includegraphics[width=\linewidth]{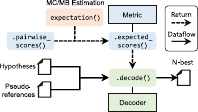}
    \caption{
    Workflow overview of \mbrs.
    \texttt{Decoder} class gets a candidate list and outputs high-quality hypotheses.
    It internally calls the \texttt{expected\_scores()} method implemented in \texttt{Metric} classes that calculate the expected scores for each hypothesis.
    The expected scores are estimated with Monte Carlo (MC) or the model-based (MB) method.
    In the figure, the colored boxes without border lines denote the functions or methods and the square boxes with border lines denote the abstract classes.
    Methods that have the same color as a class indicate that they belong to the class.
    }
    \label{fig:mbrs-workflow}
\end{figure}

We introduce \mbrs, a library of MBR decoding, which implements various metrics and algorithms with many useful features for comparisons and the development of new methods.
Figure~\ref{fig:mbrs-workflow} shows the workflow overview of \mbrs.
In the figure, each module, i.e., ``Metric'' and ``Decoder'', is easily extensible.
In addition, we carefully designed \mbrs to ensure transparency and reproducibility.
For instance, the profiler, which automatically measures how much time a code block consumes and how many times it is called, and the typed dataclass, which returns helpful metadata for analyses, are implemented.

Experiments on the WMT'22 En$\leftrightarrow$De general translation tasks demonstrated that our \mbrs can not only systematically compare various metrics and algorithms but also present the runtime statistics and supplementary information of output texts.

\section{Background}
\label{sec:background}

\subsection{Minimum Bayes risk (MBR) decoding}
The goal of MBR decoding is to find the high-quality output text from a candidate list for a given input text.
We denote input and output texts as sequences of tokens $\bm{x} \in \mathcal{V}^\ast_X$ and $\bm{y} \in \mathcal{V}^\ast_Y$, respectively.
Note that $\mathcal{V}^\ast_X$ and $\mathcal{V}^\ast_Y$ are the Kleene closures of input and output vocabularies, respectively.
Since $\bm{y}$ is searched from the output space $\mathcal{Y} \coloneqq \{ \bm{y} \mid \bm{y} \in \mathcal{V}^\ast_Y \}$, which is an infinite set, we usually find the optimal solution with some approximations.

The most commonly used strategy is maximum a posteriori (MAP) decoding using beam search.
The solution of MAP decoding $\bm{y}^{\mathrm{MAP}_\theta} \in \mathcal{Y}$ is obtained according to its output probability which is calculated by a text generation model $\theta$ as follows:
\begin{equation}
    \bm{y}^{\mathrm{MAP}_\theta} = \argmax_{\bm{y} \in \mathcal{Y}} p(\bm{y}|\bm{x}; \theta).
\end{equation}
Recent studies have demonstrated that this high-probability sequence $\bm{y}^{\mathrm{MAP}_\theta}$ is not always high-quality.
Specifically, as the beam size increases, a sequence with higher probability is obtained; however, such sequences hurt translation performance, a phenomenon known as \textit{beam search curse}.
This often results in generating pathological sequences, e.g., empty sequences, $n$-gram repetitions, and copies of the input sequence~\citep{koehn-knowles-2017-six,ott-etal-2018-analyzing,eikema-aziz-2020-map}.

In contrast, minimum Bayes risk (MBR) decoding, an alternative strategy, is known to be effective in avoiding the problems of MAP decoding~\citep{eikema-aziz-2020-map}.
It finds the high-quality text from the given candidate list $\mathcal{H} \subseteq \mathcal{Y}$ and is formulated as follows:
\begin{equation}
     \bm{y}^\mathrm{MBR_{true}} = \argmax_{\bm{h} \in \mathcal{H}} \expect_{\bm{y} \sim \Pr(\bm{y}|\bm{x})} \left[ u(\bm{h}, \bm{y}) \right],
\end{equation}
where $\Pr(\bm{y}|\bm{x})$ is the true probability of $\bm{y}$ being generated from $\bm{x}$.
$u\colon \mathcal{Y} \times \mathcal{Y} \to \mathbb{R}$ is the utility function that measures the quality or preference of generated text under the given reference text, which is defined as $\bm{h} \succeq \bm{h}' \iff u(\bm{h}, \bm{r}) \geq u(\bm{h}', \bm{r})$ where $\succeq$ denotes the preference relation.\footnote{
An evaluation metric that measures the quality of output texts, e.g., \MetricBLEU~\citep{papineni-etal-2002-bleu} or \MetricCOMET~\citep{rei-etal-2020-comet,rei-etal-2022-comet} in the machine translation task, is often employed for the utility function $u$.
}
Since the true probability $\Pr(\bm{y}|\bm{x})$ is unknown, it is replaced with the output probability calculated by a text generation model $\theta$:
\begin{equation}
     \bm{y}^{\mathrm{MBR}_\theta} = \argmax_{\bm{h} \in \mathcal{H}} \expect_{\bm{y} \sim p(\bm{y}|\bm{x}; \theta)} \left[ u(\bm{h}, \bm{y}) \right].
\end{equation}
Because enumerating all possible output texts is infeasible, the expected scores calculated by the utility function (expected utility; EU) are estimated using pseudo-references that are sampled according to the output probabilities.\footnote{
We call this calculation of the approximated expected score using pseudo-references ``expectation estimation''.
}
Let $\hat{\mathcal{R}}$ be a bag (a.k.a multiset) of sampled pseudo-references and $\mathcal{R}$ is the support set of $\hat{\mathcal{R}}$, i.e., a set of distinct elements of $\hat{\mathcal{R}}$, they are formulated as follows:
\begin{align}
    \hat{\mathcal{R}} &\coloneqq \{\bm{y}_i \in \mathcal{Y} \mid \bm{y}_i \sim p(\bm{y}|\bm{x}; \theta)\}_{i=1}^{|\hat{\mathcal{R}}|}, \\
    \mathcal{R} &\coloneqq \supp(\hat{\mathcal{R}}) = \{ \bm{y} \in \mathcal{Y} \mid m_{\hat{\mathcal{R}}}(\bm{y}) > 0 \},
\end{align}
where $m_{\hat{\mathcal{R}}}\colon \mathcal{Y} \to \mathbb{Z}_+$ is the multiplicity function\footnote{$m_{\hat{\mathcal{R}}}(\bm{y}') = 0$ where $\bm{y}' \notin \supp(\hat{\mathcal{R}})$.} which returns the number of occurrences in $\hat{\mathcal{R}}$.
Typically, the EU is estimated using the Monte Carlo method~\citep{eikema-aziz-2022-sampling} based on the empirical distribution:
\begin{align}
    p_\mathrm{MC}(\bm{r} | \bm{x}; \hat{\mathcal{R}}) &\coloneqq \frac{m_{\hat{\mathcal{R}}}(\bm{r})}{|\hat{\mathcal{R}}|}, \\
    \mu_\mathrm{MC}(\bm{h}; \hat{\mathcal{R}}) &\coloneqq \sum_{\bm{r} \in \supp(\hat{\mathcal{R}})} p_\mathrm{MC}(\bm{r}|\bm{x}; \hat{\mathcal{R}}) u(\bm{h}, \bm{r}), \\
    \bm{y}^{\mathrm{MBR}_\theta^\mathrm{MC}} &= \argmax_{\bm{h} \in \mathcal{H}} \mu_\mathrm{MC}(\bm{h}; \hat{\mathcal{R}}),
    \label{eq:mc-mbr}
\end{align}
or using the model-based method~\citep{jinnai-etal-2024-model-based} based on the output probability:
\begin{align}
    \mu_\mathrm{MB}(\bm{h}; \mathcal{R}, \theta) &\coloneqq \sum_{\bm{r} \in \mathcal{R}} p_\mathrm{MB}(\bm{r} | \bm{x}; \mathcal{R}, \theta) u(\bm{h}, \bm{r}), \\
    \bm{y}^{\mathrm{MBR}_\theta^\mathrm{MB}} &= \argmax_{\bm{h} \in \mathcal{H}} \mu_\mathrm{MB}(\bm{h}; \mathcal{R}, \theta),
    \label{eq:mb-mbr}
\end{align}
where $p_\mathrm{MB}$ is the normalized output probability over a set of pseudo-references, as follows:
\begin{equation}
    p_\mathrm{MB}(\bm{r} | \bm{x}; \mathcal{R}, \theta) \coloneqq \frac{p(\bm{r}|\bm{x}; \theta)}{\sum_{\bm{r}' \in \mathcal{R}} p(\bm{r}'|\bm{x}; \theta)}.
\end{equation}
Typically, the hypotheses themselves are regarded as the pseudo-references $\hat{\mathcal{R}}$ or $\mathcal{R}$.

\subsection{Variations in MBR decoding algorithms}
While MBR decoding can generate higher-quality texts compared with MAP decoding, the EU estimation requires the time complexity of $\mathcal{O}(|\mathcal{H}||\hat{\mathcal{R}}|)$ or $\mathcal{O}(|\mathcal{H}||\mathcal{R}|)$, which is time-consuming.
To address the issue, efficient variants of MBR decoding have been proposed.
One is the reference aggregation, which approximates the EU using an aggregated representation of pseudo-references in the feature space (RAMBR)~\citep{denero-etal-2009-fast,vamvas-sennrich-2024-linear}.
Similarly, \citet{deguchi-etal-2024-centroid} cluster pseudo-references in the embedding space and use the centroid representations for the EU estimation (CBMBR).
Another approach is hypothesis pruning, which prunes hypotheses that are not likely to be selected (PruneMBR)~\citep{cheng-vlachos-2023-faster}.
\citet{trabelsi-etal-2024-efficient} proposed probabilistic MBR (PMBR) that reduces the number of the utility function calls.
PMBR first calculates the utility scores using only sampled pairs of a hypothesis and reference instead of all pairs.
Then, it approximately computes the hypothesis--pseudo-reference utility score matrix using the low-rank matrix completion~\citep{zachariah-etal-2012-alternating}.

\subsection{Problems of existing implementations}
Currently, MBR decoding has drawn attention from research communities, and although various methods have been proposed, there is no common library that includes many of the latest studies, making it hard to compare the quality and speed systematically.
\textsc{mbr-nmt}\footnote{\textsc{mbr-nmt}: \url{https://github.com/roxot/mbr-nmt}} is the original implementation of Monte Carlo estimation~\citep{eikema-aziz-2022-sampling}, but it does not support other later methods or metrics.
\MetricCOMET~\citep{rei-etal-2020-comet,fernandes-etal-2022-quality} is known as an evaluation metric, but its framework includes a command for MBR decoding, \texttt{comet-mbr}.
It only supports \MetricCOMET as the utility function and the Monte Carlo estimation as a decoding method.
\citet{vamvas-sennrich-2024-linear} released \textsc{mbr}\footnote{\textsc{mbr}: \url{https://github.com/ZurichNLP/mbr}} which is highly integrated into huggingface's \textsc{transformers}~\citep{wolf-etal-2020-transformers}, but it only comprises their work and the vanilla MBR decoding, and it cannot be combined with other frameworks like \textsc{fairseq}~\citep{ott-etal-2019-fairseq} and black-box large language model services.

Now, to further advance MBR decoding --- a powerful technique for improving the quality of text generation --- a systematic and shared library is clearly essential for both researchers and developers.

\section{Our Library: \mbrs}
\label{sec:proposal}

Our library \mbrs is mainly implemented on Python and \textsc{PyTorch}~\citep{adam-etal-2019-pytorch}.
It finds the most suitable output from the given hypotheses.

\subsection{Main components}

\paragraph{Metrics}
Metrics are the collections of various evaluation metrics.
Basically, reference-based metrics are implemented that can be used as utility functions, but reference-free metrics are also implemented for $N$-best list reranking.
The following are the available metrics in our current repository:
\begin{itemize}
    \setlength{\itemsep}{0pt}
    \setlength{\parskip}{3pt}
    \item \MetricBLEU~\citep{papineni-etal-2002-bleu}
    \item Translation Edit Rate (\MetricTER)~\citep{snover-etal-2006-study}
    \item \MetricChrF~\citep{popovic-2015-chrf}
    \item \MetricCOMET~\citep{rei-etal-2020-comet,rei-etal-2022-comet}
    \item \MetricCometKiwi~\citep{rei-etal-2022-cometkiwi}
    \item \MetricXCOMET~\citep{guerreiro-etal-2024-xcomet}
    \item \MetricBLEURT~\citep{sellam-etal-2020-bleurt}
\end{itemize}

\paragraph{Decoders}

\begin{table*}[t]
    \centering
    \small
    \begin{tabular}{@{}lll@{}}
        \toprule
         Decoder & Available metrics & Time complexity \\
         \midrule
         \textbf{MBR}: Vanilla~\citep{eikema-aziz-2020-map,eikema-aziz-2022-sampling} & any & $\mathcal{O}(|\mathcal{H}||\hat{\mathcal{R}}|)$ \\
         \textbf{PruneMBR}: Confidence-based pruning~\citep{cheng-vlachos-2023-faster} & any & N/A \\
         \textbf{RAMBR}: Reference aggregation
         & \MetricBLEU, \MetricChrF, and \MetricCOMET & $\mathcal{O}(|\mathcal{H}| + |\hat{\mathcal{R}}|)$ \\
         \multicolumn{3}{@{}l@{}}{~~~on \MetricBLEU: Aggregate $n$-gram counts and length~\cite{denero-etal-2009-fast}} \\
         \multicolumn{3}{@{}l@{}}{~~~on \MetricChrF: Aggregate $n$-gram counts~\cite{vamvas-sennrich-2024-linear}} \\
         \multicolumn{3}{@{}l@{}}{~~~on \MetricCOMET: Aggregate sentence embeddings~\cite{vamvas-sennrich-2024-linear,deguchi-etal-2024-centroid}} \\
         \textbf{CBMBR}: Centroid-based reference aggregation~\citep{deguchi-etal-2024-centroid} & \MetricCOMET & $\mathcal{O}(|\mathcal{H}|k + |\hat{\mathcal{R}}|k)$ \\
         \textbf{PMBR}: Low-rank matrix completion~\citep{trabelsi-etal-2024-efficient} & any & $\mathcal{O}\left(\frac{|\mathcal{H}||\hat{\mathcal{R}}|}{r}\right)$ \\
         \bottomrule
    \end{tabular}    
    \caption{
    Implementation list of MBR decoding variants.
    The time complexity is based on the Monte Carlo estimation.
    Since PruneMBR dynamically decides the termination condition based on confidence that depends on the input data, we are unable to show the time complexity analytically.
    }
    \label{tab:list-efficient-decoders}
\end{table*}

\mbrs selects the most suitable output based on MBR decoding or $N$-best list reranking from the set of hypotheses $\mathcal{H}$.
When the hypotheses are reranked using reference-free metrics like \MetricCometKiwi, \mbrs performs standard $N$-best list reranking; otherwise, it performs MBR decoding.

In MBR decoding, we support not only Monte Carlo estimation~\citep{eikema-aziz-2022-sampling} but also model-based estimation~\citep{jinnai-etal-2024-model-based} by passing the log-probabilities
of pseudo-references.
These estimators can be combined with a variety of MBR decoding algorithms.
Table~\ref{tab:list-efficient-decoders} lists the implemented variants of MBR decoding.
Note that the aggregation methods in RAMBR depend on each metric.
\mbrs implements the aggregation methods for the \MetricBLEU~\citep{denero-etal-2009-fast}, \MetricChrF~\citep{vamvas-sennrich-2024-linear}, and \MetricCOMET~\citep{vamvas-sennrich-2024-linear,deguchi-etal-2024-centroid} that has been proposed so far\footnote{Other metrics, e.g., \MetricTER, \MetricXCOMET, and \MetricBLEURT, cannot aggregate the references due to their calculation~\citep{denero-etal-2009-fast,vamvas-sennrich-2024-linear,deguchi-etal-2024-centroid}.}.
The listed decoders can be combined with either the Monte Carlo or model-based estimations.

\subsection{Interfaces}
\begin{CJK}{UTF8}{ipxg}
\begin{lstlisting}[float=t, caption = An example code of MBR decoding using the \MetricCOMET metric on our \mbrs., label = lst:example-cometmbr-python]
from mbrs.metrics import MetricCOMET
from mbrs.decoders import DecoderMBR

SOURCE = "ありがとう"
HYPOTHESES = ["Thanks", "Thank you", "Thank you so much", "Thank you.", "thank you"]

metric_cfg = MetricCOMET.Config(
  model="Unbabel/wmt22-comet-da",
)
metric = MetricCOMET(metric_cfg)

decoder_cfg = DecoderMBR.Config()
decoder = DecoderMBR(decoder_cfg, metric)

output = decoder.decode(HYPOTHESES, references=HYPOTHESES, source=SOURCE, nbest=1)

print(f"Selected index: {output.idx}")
print(f"Output sentence: {output.sentence}")
print(f"Expected score: {output.score}")
\end{lstlisting}
\end{CJK}

\mbrs has two interfaces: Python application programming interface (API) and command-line interface (CLI).
Listing~\ref{lst:example-cometmbr-python} is an example code via Python API.
The detailed references of Python API and CLI are available on Read the Docs\footnote{Read the Docs: \url{https://mbrs.readthedocs.io/en/latest/index.html}}.

\subsection{Reproducibility}
\paragraph{Fixing random number seed}
Some algorithms use random numbers.
To ensure the reproducibility of experiments, \mbrs generates all random numbers from \texttt{torch.Generator} with a manual seed.

\paragraph{Dataclass/YAML-based configuration}
All metrics and decoders are configured using \texttt{dataclass}, making it more robust by using typed variables.
In CLI, the command-line arguments are automatically generated and parsed using the configuration dataclasses.
Furthermore, instead of specifying arguments directly, the YAML configuration file can also be passed via \verb|--config_path| option in CLI.

\subsection{Transparency}
\label{sec:transparency}

\paragraph{Code block profiler}

\begin{lstlisting}[float=t, caption = An example usage of our code block profiler., label = lst:example-timer-python]
>>> from mbrs import timer
>>> from mbrs.metrics import MetricBLEU
>>> HYPOTHESES = ["Thank you so much."] * 100
>>> metric = MetricBLEU(MetricBLEU.Config())
>>> scores = []
>>> for h in HYPOTHESES:
...   for r in HYPOTHESES:
...     with timer.measure("score"):
...       scores.append(metric.score(h, r))
>>> timer.aggregate().result(nsentences=1)
[{'name': 'score', 'acctime': 0.6932655478303786, 'acccalls': 10000, 'ms/call': 0.06932655478303787, 'ms/sentence': 693.2655478303786, 'calls/sentence': 10000.0}]
\end{lstlisting}

One of the key features of \mbrs is the code block profiler as shown in Listing~\ref{lst:example-timer-python}.
It can measure the elapsed time and number of calls within the code block using a context manager, i.e., \texttt{with} statement of Python.
After running the program, it automatically aggregates all profilers and reports the statistics.
This feature is helpful for identifying the bottleneck of the codes and designing new algorithms.

\paragraph{Metadata analysis}
In addition to the configuration, the outputs of the decoders are also carried by \texttt{dataclass} as shown in L23--25 of Listing~\ref{lst:example-cometmbr-python}.
At least, all decoders always return not only the output texts but also expected scores and their indices in the list of hypotheses.
This allows us to analyze where the output text came from when the hypothesis set is constructed from the multiple generation systems.
In addition, the decoders can return the reranked $N$-best lists; thus, it can show the $N$-best output texts and their expected scores.

\subsection{Extensibility}
Metrics and decoders are easily customized using the abstract classes.
By inheriting these classes, a new class can leverage the predefined common methods and needs only to implement specific methods minimally, e.g., \texttt{.score()} for metrics.
If the required methods are not implemented, an exception error will be raised.
The necessary methods can be implemented according to the type annotations and docstrings of the parent class.

\section{Experiments}
\label{sec:experiments}

\begin{table*}[t]
\centering
\small
\setlength{\tabcolsep}{3.5pt}
\begin{tabular}{@{}lrr rrrrr rrrrr @{}}
\toprule
&&& \multicolumn{5}{c}{En--De} & \multicolumn{5}{c}{De--En} \\
 \cmidrule(lr){4-8} \cmidrule(l){9-13} 
Decoding & Utility & Est. & \MetricBLEU & \MetricChrF & \MetricCOMET & \MetricBLEURT & \MetricKiwi & \MetricBLEU & \MetricChrF & \MetricCOMET & \MetricBLEURT & \MetricKiwi  \\
\midrule
MAP$_\epsilon$ & -- & -- & 23.8 & 50.1 & 75.6 & 54.9 & 74.2 & 26.0 & 50.8 & 78.4 & 62.0 & 75.4 \\
MAP$_\text{beam}$ & -- & -- & 25.1 & 52.8 & 77.3 & 56.5 & 76.0 & 27.3 & 52.5 & 79.2 & 62.7 & 76.2 \\
QE & \MetricKiwi & -- & 22.5 & 51.8 & 82.1 & 58.4 & 83.6 & 23.8 & 50.7 & 82.2 & 63.3 & 81.9 \\
\midrule
MBR & \MetricBLEU & MC & 25.6 & 52.9 & 75.2 & 54.8 & 73.7 & 27.2 & 52.1 & 78.5 & 62.0 & 75.3 \\
 & & MB & 24.2 & 50.5 & 75.5 & 54.8 & 74.1 & 26.4 & 51.1 & 78.5 & 61.9 & 75.3 \\
 & \MetricChrF & MC & 24.4 & 54.2 & 76.3 & 56.4 & 75.1 & 27.5 & 53.4 & 79.0 & 62.8 & 76.0 \\
 & & MB & 24.4 & 51.2 & 76.0 & 55.7 & 74.8 & 26.8 & 51.9 & 78.8 & 62.5 & 76.0 \\
 & \MetricCOMET & MC & 24.1 & 52.8 & 83.9 & 58.7 & 79.5 & 25.7 & 51.9 & 83.0 & 63.7 & 78.5 \\
 & & MB & 24.1 & 51.1 & 81.0 & 57.1 & 77.5 & 26.2 & 51.4 & 81.6 & 63.2 & 77.6 \\
 & \MetricBLEURT & MC & OOT & OOT & OOT & OOT & OOT & OOT & OOT & OOT & OOT & OOT \\
 & & MB & OOT & OOT & OOT & OOT & OOT & OOT & OOT & OOT & OOT & OOT \\
 \cmidrule(l){2-13}
RAMBR & \MetricBLEU & MC & 25.6 & 52.9 & 75.2 & 54.9 & 73.8 & 27.3 & 52.1 & 78.6 & 62.1 & 75.4 \\
 & & MB & 24.2 & 50.5 & 75.5 & 54.8 & 74.1 & 26.4 & 51.1 & 78.5 & 62.0 & 75.4 \\
 & \MetricChrF & MC & 24.3 & 54.2 & 76.1 & 56.3 & 75.0 & 27.4 & 53.4 & 79.0 & 62.9 & 76.0 \\
 & & MB & 24.4 & 51.1 & 75.8 & 55.6 & 74.7 & 26.9 & 51.9 & 78.9 & 62.6 & 76.0 \\
 & \MetricCOMET & MC & 23.9 & 52.9 & 83.0 & 58.1 & 79.1 & 26.0 & 51.9 & 82.3 & 63.5 & 78.0 \\
 & & MB & 24.3 & 51.0 & 80.1 & 56.6 & 77.0 & 26.2 & 51.4 & 81.0 & 63.1 & 77.2 \\
 \cmidrule(l){2-13}
CBMBR & \MetricCOMET & MC & 23.2 & 52.4 & 83.8 & 58.3 & 79.2 & 25.1 & 51.4 & 83.0 & 63.3 & 78.3 \\
 & & MB & 22.6 & 51.6 & 82.2 & 57.4 & 77.8 & 24.2 & 50.4 & 81.9 & 62.6 & 77.4 \\
 \cmidrule(l){2-13}
Pruning & \MetricBLEU & MC & 25.7 & 52.9 & 75.0 & 54.6 & 73.6 & 27.1 & 52.0 & 78.5 & 61.9 & 75.3 \\
 & & MB & 24.3 & 52.0 & 75.5 & 54.9 & 74.2 & 26.2 & 51.4 & 78.6 & 61.9 & 75.4 \\
 & \MetricChrF & MC & 24.5 & 54.3 & 76.2 & 56.3 & 75.1 & 27.4 & 53.5 & 78.9 & 62.8 & 75.9 \\
 & & MB & 24.6 & 52.8 & 76.1 & 55.9 & 75.0 & 26.8 & 52.3 & 78.8 & 62.4 & 75.9 \\
 & \MetricCOMET & MC & 24.0 & 52.8 & 83.9 & 58.7 & 79.5 & 25.8 & 51.9 & 83.0 & 63.7 & 78.4 \\
 & & MB & 24.4 & 52.5 & 82.1 & 57.9 & 78.4 & 26.3 & 51.9 & 81.6 & 63.3 & 77.6 \\
 & \MetricBLEURT & MC & 22.7 & 51.7 & 77.6 & 63.4 & 75.6 & 24.9 & 51.3 & 79.4 & 65.1 & 76.4 \\
 & & MB & 24.1 & 52.2 & 77.3 & 59.4 & 75.9 & 26.0 & 51.5 & 79.1 & 63.5 & 76.1 \\
 \cmidrule(l){2-13}
PMBR & \MetricBLEU & MC & 25.0 & 52.2 & 74.2 & 54.4 & 72.8 & 26.4 & 51.4 & 78.2 & 61.8 & 75.0 \\
 & & MB & 24.9 & 52.0 & 73.7 & 54.2 & 72.3 & 26.4 & 51.5 & 78.0 & 61.6 & 74.7 \\
 & \MetricChrF & MC & 23.0 & 52.9 & 74.3 & 55.0 & 73.1 & 26.0 & 52.2 & 78.3 & 62.1 & 75.1 \\
 & & MB & 22.9 & 52.7 & 74.1 & 54.8 & 72.8 & 25.7 & 52.1 & 78.1 & 61.9 & 75.0 \\
 & \MetricCOMET & MC & 23.9 & 52.7 & 83.8 & 58.7 & 79.5 & 25.9 & 51.9 & 82.8 & 63.6 & 78.3 \\
 & & MB & 23.6 & 51.8 & 82.6 & 57.7 & 78.6 & 25.7 & 51.5 & 82.1 & 63.5 & 77.9 \\
 & \MetricBLEURT & MC & 22.7 & 51.8 & 77.7 & 63.3 & 75.7 & 25.0 & 51.4 & 79.4 & 65.1 & 76.4 \\
 & & MB & 23.1 & 51.5 & 77.2 & 60.8 & 75.8 & 25.5 & 51.4 & 79.2 & 64.1 & 76.3 \\
 \midrule
Oracle & \MetricBLEU & -- & 44.4 & 62.8 & 76.1 & 59.5 & 71.6 & 46.4 & 63.4 & 80.9 & 66.6 & 74.3 \\
 & \MetricChrF & -- & 39.5 & 66.6 & 77.7 & 62.0 & 72.9 & 43.5 & 66.0 & 81.3 & 67.4 & 74.8 \\
 & \MetricCOMET & -- & 30.2 & 58.0 & 87.0 & 64.1 & 80.4 & 34.4 & 58.3 & 86.4 & 68.5 & 78.9 \\
 & \MetricBLEURT & -- & 31.0 & 58.2 & 79.7 & 71.0 & 75.0 & 35.7 & 59.0 & 82.5 & 72.6 & 76.3 \\
\bottomrule
\end{tabular}
\caption{
Experimental results on the WMT'22 general MT task in En--De and De--En.
}
\label{tab:results-ende}
\end{table*}

\subsection{Setup}
\label{sec:setup}

Using our \mbrs, we conducted translation experiments on the WMT'22 general translation task~\citep{kocmi-etal-2022-findings} in En--De and De--En, and evaluated the translation quality and speed.
We compared various MBR decoding methods: vanilla~\citep{eikema-aziz-2020-map,eikema-aziz-2022-sampling}, reference aggregation~\citep{denero-etal-2009-fast,vamvas-sennrich-2024-linear}, centroid-based aggregation~\citep{deguchi-etal-2024-centroid}, confidence-based pruning~\citep{cheng-vlachos-2023-faster}, and probabilistic MBR~\citep{trabelsi-etal-2024-efficient}.
In addition to MBR decoding, we compared other decoding methods, MAP decoding (MAP) and $N$-best reranking using a quality estimation model (QE).
We also measured the upper bound of the translation quality (Oracle) by selecting translations that maximize the evaluation metric using the reference translations.

Translation candidates were generated using  M2M100\footnote{\url{https://huggingface.co/facebook/m2m100_418M}}~\citep{fan-etal-2021-beyond}.
For QE reranking, MBR decoding, and oracle evaluation, we sampled 1,024 translations using epsilon sampling with $\epsilon=0.02$.
We used the same collection for hypotheses and pseudo-references.
In MAP decoding, we compared the same hypotheses as MBR decoding (MAP$_\epsilon$) and the higher-probability hypotheses generated by beam search with a beam size of 256 (MAP$_\text{beam}$).

We evaluated the translation quality on \MetricBLEU~\citep{papineni-etal-2002-bleu}, \MetricChrF~\citep{popovic-2015-chrf}, \MetricCOMET~\citep{rei-etal-2022-comet}, \MetricBLEURT~\citep{sellam-etal-2020-bleurt}, and \MetricCometKiwi (\MetricKiwi)~\citep{rei-etal-2022-cometkiwi}.
We employed \MetricKiwi for QE reranking, and other metrics for the utility functions of MBR decoding and the oracle.
The detailed setup is described in Appendix~\ref{sec:detailed-setup}.

\subsection{Results}

\begin{table}[t]
    \centering
    \small
    \begin{tabular}{@{}lrrrr@{}}
        \toprule
        Step & MBR & PruneMBR & RAMBR & PMBR \\
        \midrule
        Prune & -- & 950.7 & -- & -- \\
        Aggregate  & -- & -- & 494.3 & -- \\
        ALS & -- & -- & -- & 159.3 \\
        Utility & 15789.8 & 12158.0 & 104.3 & 2268.6 \\
        \midrule
        E2E & 16076.0 & 11206.8 & 605.6 & 2704.2 \\
        \bottomrule
    \end{tabular}
    \caption{The execution time in the WMT'22 En--De translation task when \MetricBLEU is employed as the utility function.}
    \label{tab:results-speed-bleu}
\end{table}

\begin{table}[t]
    \centering
    \small
    \tabcolsep 3pt
    \begin{tabular}{@{}lrrrrrr@{}}
        \toprule
                & & & Prune- & RA-& CB-& \\
        Step & QE & MBR & MBR & MBR & MBR & PMBR \\
        \midrule
        Rerank & 304.0 & -- & -- & -- & -- & --\\
        Encode & -- & 316.2 & 310.4 & 325.0 & 324.0 & 319.2 \\ 
        Prune & -- & -- & 5.6 & -- & -- & --\\
        Aggregate &  -- & -- & --  & 0.0 & -- &  -- \\
        Clustering & -- & -- & --  & -- & 21.2 &  -- \\
        ALS & -- & -- & -- & -- & -- & 5.9 \\
        Utility & -- & 277.5 & 45.3 & 0.3 & 11.4 &  21.4\\
        \midrule
        E2E & 522.3 & 618.2 & 361.6 & 325.9 & 363.9 & 567.3\\
        \bottomrule
    \end{tabular}
    \caption{The execution time in the WMT'22 En--De translation task when \MetricCOMET is employed as the utility function.}
    \label{tab:results-speed-comet}
\end{table}

\paragraph{Quality of decoded texts}
Table~\ref{tab:results-ende} shows the experimental results of the translation quality.
In the table, ``Est.'' indicates the estimation of expected utility.
``MC'' and ``MB'' denote the Monte Carlo and model-based estimations, respectively.
Due to our computational limitations, the jobs exceeding 100 hours are noted as ``OOT''.
The results show that our \mbrs works in various combinations of metrics and decoding methods, and improve the quality of texts compared with both MAP$_\epsilon$ and MAP$_\text{beam}$.
We confirmed that several approximated algorithms also work well, e.g., RAMBR, CBMBR, PruneMBR, and PMBR improved the \MetricCOMET scores when the \MetricCOMET was used as the utility function.

To summarize, \mbrs can compare various metrics and decoding methods systematically.

\paragraph{Execution time}
Table~\ref{tab:results-speed-bleu} and~\ref{tab:results-speed-comet} show the execution time in milliseconds (msec) per sentence when we used \MetricBLEU and \MetricCOMET as the utility functions, respectively, in the WMT'22 En--De translation.
``QE'' in Table~\ref{tab:results-speed-comet} shows the reference time of QE reranking with \MetricKiwi.
In the tables, ``Step'' indicates each step of decoding, and ``E2E'' reports the end-to-end total time including miscellaneous processes.
Both tables demonstrate that alternative algorithms of the vanilla MBR decoding improved the speed.
The additional results with other utility functions are shown in Appendix~\ref{sec:additional-experiments}.

\mbrs can measure the time spent for each profiling code block as shown in the tables, it is helpful for identifying bottlenecks and developing new algorithms.

\paragraph{Distribution of the expected utility}
\begin{figure}[t]
    \centering
    \includegraphics[width=\linewidth]{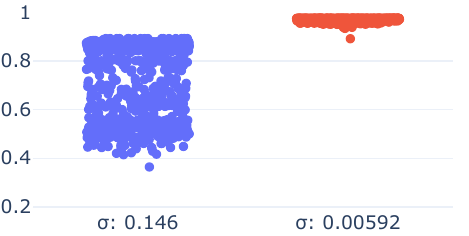}
    \caption{The distribution of expected utility scores in the set of hypotheses.
    The left and right ones show the examples that have the maximum and minimum variance of the expected utility in the test set, respectively.
    }
    \label{fig:variance-eu}
\end{figure}

\begin{figure}[t]
    \centering
    \includegraphics[width=\linewidth]{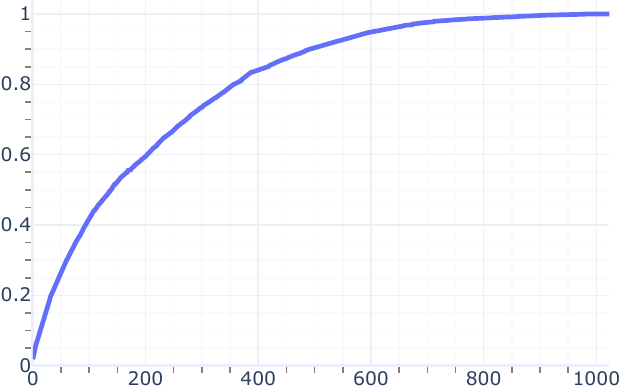}
    \caption{The empirical cumulative distribution of ranks selected by MBR decoding from descending rankings of output probabilities.
    }
    \label{fig:accumurate}
\end{figure}

As described in Section~\ref{sec:transparency}, \mbrs can present the additional information of the output texts.
We visualized the distribution of expected utility scores in the hypothesis set by using them.
Figure~\ref{fig:variance-eu} compares the distribution of the expected utility scores between the test cases that have the maximum and minimum variance in the test set of the WMT'22 En--De translation task.
We observed that there are cases in which the expected utility varies significantly depending on each hypothesis and cases in which it does not vary much.
The latter result suggests that the $N$-best outputs of MBR decoding are sometimes similar to each other.

\mbrs enables us this analysis by returning the metadata of the output texts.

\paragraph{Visualize the decision of MBR decoding}

\mbrs also returns which indices in the hypotheses are selected by MBR decoding.
Figure~\ref{fig:accumurate} shows the empirical cumulative distribution of ranks selected by MBR decoding from descending rankings of output probabilities in the WMT'22 En--De translation task.
In the figure, the x-axis indicates the 0-indexed rank of output probabilities in the hypothesis set, i.e., 0 is the highest probability in the hypotheses set, and the y-axis indicates the ratio of the number of selected by MBR decoding when using \MetricCOMET as the utility function.

From the results, by using \mbrs, we could observe that MBR decoding does not always select high-probability texts, i.e., high-quality texts and high-probability texts are different, as mentioned by~\citet{freitag-etal-2022-high}.

\section{Conclusion}
\label{sec:conclusion}

This paper describes \mbrs, a library for MBR decoding, which implements various metrics and decoding methods.
It is designed to ensure transparency, reproducibility, and extensibility for researchers and developers.
We will continue to implement the latest metrics and decoding methods\footnote{
This paper describes \mbrs up to the version \texttt{v0.1.2}.
The latest version, \texttt{v0.1.3}, now implements diverse MBR~\citep{jinnai-etal-2024-generating}.
}.
Furthermore, we will support evaluation metrics other than those used in the machine translation task.
We hope our \mbrs will be used as a tool to further improve the quality of text generation models.

\section*{Limitations}
We measured the execution times only in a single run; thus, the speed may differ when different computer architectures are used.

Currently, our repository mainly implements metrics for translation tasks.
Implementing evaluation metrics other than the translation task is a future work.

\section*{Ethics Statement and Broader Impact}
MBR decoding selects output texts from a set of a candidate list generated by text generation models; hence, if the systems generate toxic text, it may be selected depending on the utility functions.
In addition, if a utility function 
is biased and not aligned with human preferences,
MBR decoding is more likely to generate biased texts.
Since MBR decoding reflects human preferences in the output texts through the utility function, it must be carefully designed to avoid generating harmful texts.

\section*{Acknowledgments}
This work was supported by JSPS KAKENHI Grant Number 
JP21H05054 and
JP23H03458.

\bibliography{anthology,custom}

\clearpage
\appendix

\section{License}
\mbrs is published under the MIT license.
Our implementation calls the \MetricCOMET and \MetricBLEURT models, licensed under the 
Apache-2.0 license, and the \MetricCometKiwi and \MetricXCOMET models, licensed under the CC BY-NC-SA 4.0 license.
We used the test sets of WMT'22 general translation tasks published under the following policy: ``The data released for the WMT General MT task can be freely used for research purposes''.

\section{Links to Our Project}
\label{sec:links}
The following are links to our project pages:
\begin{itemize}[left=5pt]
    \setlength{\itemsep}{0pt}
    \setlength{\parskip}{3pt}
    \setlength{\itemindent}{0pt}
	\setlength{\labelsep}{5pt}
    \item GitHub: \small\url{https://github.com/naist-nlp/mbrs}\normalsize
    \item HP: \small\url{https://naist-nlp.github.io/mbrs-web}\normalsize
    \item Docs: \small\url{https://mbrs.readthedocs.io/en/latest}\normalsize
    \item YouTube: \small\url{https://youtu.be/4qeHpg4PTn0}\normalsize
\end{itemize}

\section{Detailed Experimental Setup}
\label{sec:detailed-setup}

In CBMBR, we set the number of centroids to 64.
The reduction factor of PMBR was set to 8 and we set $\alpha=0.99$ in PruneMBR.

We measured the speeds on an NVIDIA\regmark{} RTX\trademark{} 6000 Ada GPU with a batch size of 256 and the half-precision computation for \MetricCOMET, \MetricKiwi, and \MetricBLEURT metrics, and on the 32 core Intel\regmark{} Xeon\regmark{} Platinum 8160 CPUs for \MetricBLEU and \MetricChrF metrics.

Table~\ref{tab:metric-model-names} lists the model names that we used for the evaluation metrics.

\begin{table}[t]
    \centering
    \begin{tabular}{@{}ll@{}}
    \toprule
        Metric & Model \\
        \midrule
        \MetricCOMET & \texttt{Unbabel/wmt22-comet-da} \\
        \MetricKiwi & \texttt{Unbabel/wmt22-cometkiwi-da} \\
        \MetricBLEURT & \texttt{BLEURT-20-D12} \\
        \bottomrule
    \end{tabular}
    \caption{Model list of evaluation metrics that we used in the experiments.}
    \label{tab:metric-model-names}
\end{table}

\section{Additional Experiments}
\label{sec:additional-experiments}

\subsection{Execution time on the other metrics}

Table~\ref{tab:results-speed-chrf} and~\ref{tab:results-speed-bleurt} show the execution times of MBR decoding in the WMT'22 En--De translation task when \MetricChrF and \MetricBLEURT are used as the utility functions, respectively.

\subsection{Additional translation experiments on other language directions}
Table~\ref{tab:results-enzh-scores} and \ref{tab:results-enja-scores} show the comparisons of the translation quality in the WMT'22 En$\leftrightarrow$Zh and En$\leftrightarrow$Ja general translation tasks, respectively.
The experimental setup is the same as the experiments of the WMT'22 En$\leftrightarrow$De translation tasks described in Section~\ref{sec:setup}.

\begin{table}[t]
    \centering
    \small
    \begin{tabular}{@{}lrrrr@{}}
        \toprule
        Step & MBR & PruneMBR & RAMBR & PMBR \\
        \midrule
        Prune & -- & 958.4 & -- & -- \\
        Aggregate  & -- & -- & 1090.8  & --\\
        ALS & -- & -- & --  & 96.2\\
        Utility & 36540.7 & 10792.9 & 539.7 & 4694.3\\
        \midrule
        E2E & 36695.3 & 11751.7 & 1642.5 & 5019.6\\
        \bottomrule
    \end{tabular}
    \caption{The execution time when \MetricChrF is employed as the utility function.}
    \label{tab:results-speed-chrf}
\end{table}

\begin{table}[t]
    \centering
    \small
    \begin{tabular}{@{}lrrr@{}}
        \toprule
        Step & MBR & PruneMBR & PMBR \\
        \midrule
        Prune & -- & 322.0& -- \\
        Aggregate  & -- & -- & -- \\
        ALS & -- & -- & 5.5\\
        Utility & OOT & 3026.8 & 38647.1\\
        \midrule
        E2E & OOT & 3349.0 & 38797.8\\
        \bottomrule
    \end{tabular}
    \caption{The execution time when \MetricBLEURT is employed as the utility function.}
    \label{tab:results-speed-bleurt}
\end{table}

\begin{table*}[t]
\centering
\small
\tabcolsep 3.5pt
\begin{tabular}{@{}lrr rrrrr rrrrr@{}}
\toprule
&&& \multicolumn{5}{c}{En--Zh} & \multicolumn{5}{c}{Zh--En} \\
 \cmidrule(lr){4-8} \cmidrule(l){9-13} 
Decoding & Utility & Est. & \MetricBLEU & \MetricChrF & \MetricCOMET & \MetricBLEURT & \MetricKiwi & \MetricBLEU & \MetricChrF & \MetricCOMET & \MetricBLEURT & \MetricKiwi  \\
\midrule

MAP$_\epsilon$ & -- & -- & 25.1 & 25.0 & 76.4 & 53.0 & 73.3 & 12.1 & 35.0 & 67.2 & 49.5 & 65.4 \\
MAP$_\text{beam}$ & -- & -- & 28.2 & 27.0 & 77.4 & 53.6 & 74.2 & 16.1 & 43.0 & 70.1 & 51.7 & 68.4 \\
QE & \MetricKiwi &  & 25.8 & 25.1 & 83.0 & 55.7 & 82.8 & 15.4 & 46.6 & 76.9 & 55.0 & 77.7 \\
\midrule
MBR & \MetricBLEU & MC & 29.8 & 27.8 & 77.3 & 53.8 & 74.2 & 17.2 & 46.8 & 71.9 & 51.8 & 69.4 \\
 & & MB & 25.6 & 25.5 & 76.5 & 53.2 & 73.4 & 12.4 & 35.4 & 67.4 & 49.7 & 65.7 \\
 & \MetricChrF & MC & 29.5 & 28.5 & 77.6 & 53.8 & 74.4 & 17.3 & 48.9 & 72.8 & 53.1 & 70.4 \\
 & & MB & 25.4 & 25.8 & 76.8 & 53.4 & 73.7 & 12.3 & 35.9 & 67.9 & 50.5 & 66.2 \\
 &  & MC & 28.1 & 26.5 & 84.1 & 55.4 & 78.7 & 15.8 & 46.7 & 77.3 & 54.1 & 73.0 \\
 & & MB & 25.5 & 25.7 & 81.4 & 54.5 & 76.6 & 12.3 & 35.6 & 69.9 & 50.8 & 67.4 \\
 & \MetricBLEURT & MC & OOT & OOT & OOT & OOT & OOT & OOT & OOT & OOT & OOT & OOT \\
 & & MB & OOT & OOT & OOT & OOT & OOT & OOT & OOT & OOT & OOT & OOT \\
 \cmidrule(l){2-13}
RAMBR & \MetricBLEU & MC & 29.8 & 27.7 & 77.3 & 53.7 & 74.2 & 17.2 & 46.8 & 72.1 & 52.0 & 69.6 \\
 & & MB & 25.4 & 25.4 & 76.4 & 53.2 & 73.3 & 12.0 & 35.3 & 67.3 & 49.8 & 65.6 \\
 & \MetricChrF & MC & 29.4 & 28.6 & 77.5 & 53.5 & 74.4 & 17.1 & 48.7 & 72.8 & 53.1 & 70.4 \\
 & & MB & 25.1 & 25.7 & 76.6 & 53.0 & 73.6 & 12.2 & 35.8 & 67.8 & 50.5 & 66.2 \\
 & \MetricCOMET & MC & 28.7 & 26.9 & 83.2 & 55.4 & 78.0 & 16.0 & 46.9 & 76.6 & 53.9 & 72.6 \\
 & & MB & 25.6 & 25.7 & 80.5 & 54.3 & 76.0 & 12.3 & 35.7 & 69.6 & 50.7 & 67.2 \\
 \cmidrule(l){2-13}
CBMBR & \MetricCOMET & MC & 27.6 & 26.0 & 84.0 & 55.2 & 78.5 & 15.7 & 46.5 & 77.2 & 53.7 & 72.8 \\
 & & MB & 26.9 & 25.6 & 82.8 & 54.4 & 77.5 & 14.5 & 44.5 & 75.0 & 52.6 & 71.1 \\
 \cmidrule(l){2-13}
Pruning & \MetricBLEU & MC & 29.8 & 27.8 & 77.3 & 53.8 & 74.2 & 17.2 & 46.8 & 72.0 & 51.8 & 69.4 \\
 & & MB & 28.5 & 27.1 & 77.3 & 53.7 & 74.1 & 15.5 & 44.1 & 71.2 & 51.4 & 68.9 \\
 & \MetricChrF & MC & 29.5 & 28.5 & 77.6 & 53.8 & 74.5 & 17.2 & 48.8 & 72.8 & 53.2 & 70.5 \\
 & & MB & 28.8 & 27.4 & 77.6 & 53.9 & 74.4 & 15.9 & 45.2 & 71.7 & 52.2 & 69.4 \\
 & \MetricCOMET & MC & 28.1 & 26.5 & 84.1 & 55.5 & 78.6 & 15.9 & 46.8 & 77.3 & 54.1 & 72.9 \\
 & & MB & 28.9 & 27.3 & 82.3 & 55.1 & 77.4 & 15.2 & 44.3 & 74.4 & 52.7 & 70.9 \\
 & \MetricBLEURT & MC & 27.3 & 25.9 & 78.0 & 58.7 & 75.0 & 15.7 & 46.5 & 73.6 & 56.7 & 71.3 \\
 & & MB & 28.3 & 26.9 & 77.7 & 55.7 & 74.6 & 15.1 & 44.2 & 72.0 & 53.7 & 69.7 \\
 \cmidrule(l){2-13}
PMBR & \MetricBLEU & MC & 29.0 & 27.1 & 77.0 & 53.7 & 73.9 & 16.9 & 46.5 & 71.8 & 51.7 & 69.2 \\
 & & MB & 28.9 & 27.0 & 76.7 & 53.2 & 73.7 & 16.5 & 45.8 & 71.4 & 51.4 & 68.8 \\
 & \MetricChrF & MC & 29.0 & 27.9 & 77.3 & 53.7 & 74.2 & 16.0 & 47.6 & 72.2 & 52.5 & 69.8 \\
 & & MB & 28.7 & 27.8 & 76.9 & 53.1 & 73.8 & 15.9 & 47.4 & 72.0 & 52.4 & 69.6 \\
 & \MetricCOMET & MC & 28.3 & 26.6 & 83.9 & 55.5 & 78.5 & 15.9 & 46.7 & 77.1 & 54.0 & 72.9 \\
 & & MB & 27.0 & 26.1 & 82.5 & 54.9 & 77.4 & 14.1 & 41.5 & 73.5 & 52.3 & 70.0 \\
 & \MetricBLEURT & MC & 27.6 & 26.2 & 78.0 & 58.6 & 74.9 & 15.5 & 46.4 & 73.5 & 56.6 & 71.2 \\
 & & MB & 27.7 & 26.3 & 77.6 & 56.9 & 74.5 & 14.7 & 44.1 & 72.0 & 54.4 & 69.6 \\
\midrule
Oracle & \MetricBLEU & -- & 45.5 & 41.9 & 80.9 & 61.4 & 74.2 & 31.9 & 54.9 & 74.4 & 56.3 & 69.0 \\
 & \MetricChrF & -- & 45.6 & 42.8 & 81.0 & 61.3 & 74.1 & 28.2 & 58.2 & 75.5 & 57.8 & 70.0 \\
 & \MetricCOMET & -- & 35.4 & 33.4 & 87.6 & 61.8 & 79.3 & 20.5 & 50.8 & 81.2 & 58.9 & 73.9 \\
 & \MetricBLEURT & -- & 36.3 & 34.2 & 81.9 & 68.6 & 75.6 & 21.3 & 50.9 & 76.6 & 63.8 & 71.6 \\
\bottomrule
\end{tabular}
\caption{Comparisons of the translation quality in the WMT'22 En$\leftrightarrow$Zh general translation tasks.}
\label{tab:results-enzh-scores}
\end{table*}

\begin{table*}[t]
\centering
\small
\tabcolsep 3.5pt
\begin{tabular}{@{}lrr rrrrr rrrrr@{}}
\toprule
&&& \multicolumn{5}{c}{En--Ja} & \multicolumn{5}{c}{Ja--En} \\
 \cmidrule(lr){4-8} \cmidrule(l){9-13}
Decoding & Utility & Est. & \MetricBLEU & \MetricChrF & \MetricCOMET & \MetricBLEURT & \MetricKiwi & \MetricBLEU & \MetricChrF & \MetricCOMET & \MetricBLEURT & \MetricKiwi  \\
\midrule
MAP$_\epsilon$ & -- & -- & 15.1 & 22.6 & 79.2 & 46.1 & 77.4 & 9.0 & 29.2 & 68.0 & 46.5 & 68.5 \\
MAP$_\text{beam}$ & -- & -- & 14.6 & 22.3 & 78.6 & 46.0 & 77.4 & 10.9 & 33.9 & 69.7 & 48.5 & 70.5 \\
QE & \MetricKiwi &  & 16.3 & 26.1 & 86.4 & 50.8 & 86.7 & 10.8 & 37.0 & 76.3 & 52.3 & 80.1 \\
\midrule
MBR & \MetricBLEU & MC & 17.9 & 26.2 & 81.5 & 48.2 & 80.0 & 11.7 & 35.5 & 70.1 & 47.4 & 69.9 \\
 & & MB & 15.6 & 23.1 & 79.9 & 46.8 & 78.0 & 9.4 & 29.6 & 68.1 & 46.3 & 68.5 \\
 & \MetricChrF & MC & 16.8 & 26.7 & 81.9 & 49.1 & 80.9 & 12.0 & 38.4 & 70.9 & 49.5 & 71.1 \\
 & & MB & 15.6 & 23.4 & 80.1 & 47.3 & 78.6 & 9.8 & 31.0 & 68.8 & 47.7 & 69.5 \\
 & \MetricCOMET & MC & 16.6 & 26.0 & 87.9 & 50.5 & 83.5 & 10.5 & 36.0 & 76.7 & 50.4 & 74.0 \\
 & & MB & 15.7 & 23.6 & 84.0 & 48.7 & 80.5 & 9.6 & 30.4 & 72.2 & 48.4 & 71.2 \\
 & \MetricBLEURT & MC & OOT & OOT & OOT & OOT & OOT & OOT & OOT & OOT & OOT & OOT \\
 & & MB & OOT & OOT & OOT & OOT & OOT & OOT & OOT & OOT & OOT & OOT \\
 \cmidrule(l){2-13}
RAMBR & \MetricBLEU & MC & 17.7 & 26.0 & 81.5 & 48.0 & 80.0 & 11.8 & 35.6 & 70.1 & 47.8 & 70.0 \\
 & & MB & 15.5 & 23.0 & 79.7 & 46.6 & 77.9 & 9.4 & 29.9 & 68.1 & 46.7 & 68.7 \\
 & \MetricChrF & MC & 16.8 & 26.8 & 81.8 & 49.1 & 80.8 & 11.8 & 38.4 & 71.0 & 49.5 & 71.2 \\
 & & MB & 15.3 & 23.4 & 80.0 & 47.3 & 78.6 & 9.8 & 31.1 & 68.8 & 47.9 & 69.5 \\
 & \MetricCOMET & MC & 16.9 & 26.2 & 87.3 & 50.5 & 83.4 & 10.8 & 36.2 & 75.8 & 50.1 & 73.4 \\
 & & MB & 15.9 & 23.6 & 83.3 & 48.3 & 80.1 & 9.4 & 30.3 & 71.5 & 48.1 & 70.6 \\
 \cmidrule(l){2-13}
CBMBR & \MetricCOMET & MC & 16.5 & 26.0 & 87.8 & 50.1 & 83.3 & 10.0 & 35.4 & 76.6 & 50.1 & 73.5 \\
 & & MB & 15.7 & 24.9 & 86.4 & 49.0 & 82.3 & 9.3 & 33.8 & 74.5 & 48.8 & 72.1 \\
 \cmidrule(l){2-13}
Pruning & \MetricBLEU & MC & 17.9 & 26.2 & 81.5 & 48.2 & 80.0 & 11.7 & 35.4 & 70.0 & 47.4 & 69.8 \\
 & & MB & 16.5 & 25.1 & 81.3 & 47.5 & 79.4 & 10.1 & 33.1 & 69.4 & 46.9 & 69.3 \\
 & \MetricChrF & MC & 16.9 & 26.8 & 81.8 & 49.1 & 80.9 & 11.9 & 38.3 & 70.8 & 49.5 & 71.0 \\
 & & MB & 16.6 & 25.4 & 81.5 & 48.0 & 79.9 & 10.8 & 34.9 & 70.2 & 48.4 & 70.3 \\
 & \MetricCOMET & MC & 16.5 & 26.0 & 87.9 & 50.4 & 83.6 & 10.4 & 35.9 & 76.6 & 50.3 & 73.9 \\
 & & MB & 17.0 & 25.7 & 86.1 & 49.7 & 82.3 & 10.0 & 34.1 & 73.9 & 49.2 & 72.2 \\
 & \MetricBLEURT & MC & 15.8 & 25.5 & 82.5 & 53.9 & 80.6 & 10.2 & 35.2 & 71.3 & 53.2 & 71.4 \\
 & & MB & 16.2 & 25.0 & 81.9 & 50.0 & 80.0 & 9.9 & 33.2 & 70.4 & 50.0 & 70.4 \\
 \cmidrule(l){2-13}
PMBR & \MetricBLEU & MC & 17.4 & 25.7 & 81.1 & 48.1 & 79.7 & 11.2 & 34.9 & 69.9 & 47.4 & 69.8 \\
 & & MB & 17.3 & 25.7 & 81.2 & 47.8 & 79.6 & 11.1 & 34.6 & 69.4 & 47.1 & 69.2 \\
 & \MetricChrF & MC & 16.4 & 26.1 & 81.3 & 48.5 & 80.2 & 11.1 & 37.2 & 70.4 & 49.0 & 70.6 \\
 & & MB & 16.3 & 26.0 & 81.1 & 48.2 & 79.9 & 11.2 & 37.2 & 70.1 & 48.7 & 70.2 \\
 & \MetricCOMET & MC & 16.7 & 26.0 & 87.7 & 50.3 & 83.4 & 10.5 & 35.8 & 76.5 & 50.3 & 73.8 \\
 & & MB & 16.0 & 24.7 & 85.9 & 49.4 & 82.1 & 10.0 & 32.6 & 74.0 & 49.1 & 72.1 \\
 & \MetricBLEURT & MC & 15.8 & 25.5 & 82.5 & 53.7 & 80.6 & 10.1 & 35.2 & 71.5 & 53.2 & 71.6 \\
 & & MB & 16.0 & 24.7 & 82.0 & 51.2 & 79.9 & 9.6 & 33.5 & 70.5 & 51.1 & 70.7 \\
 \midrule
Oracle & \MetricBLEU & -- & 34.7 & 39.0 & 83.4 & 54.4 & 78.7 & 28.0 & 46.2 & 73.2 & 53.1 & 69.5 \\
 & \MetricChrF & -- & 33.2 & 41.4 & 84.2 & 56.3 & 79.2 & 24.0 & 50.4 & 74.5 & 55.2 & 70.8 \\
 & \MetricCOMET & -- & 22.4 & 31.6 & 90.6 & 56.8 & 83.6 & 15.9 & 41.2 & 81.9 & 56.3 & 75.5 \\
 & \MetricBLEURT & -- & 23.5 & 33.0 & 85.3 & 64.3 & 80.4 & 16.3 & 41.6 & 75.9 & 62.2 & 72.8 \\
\bottomrule
\end{tabular}
\caption{Comparisons of the translation quality in the WMT'22 En$\leftrightarrow$Ja general translation tasks.}
\label{tab:results-enja-scores}
\end{table*}

\end{document}